\documentclass[sigconf]{acmart}
\usepackage{subfigure}
\usepackage{tabularx}
\usepackage{makecell}

\usepackage{multirow}
\AtBeginDocument{%
  }

\copyrightyear{2024}
\acmYear{2024}
\setcopyright{rightsretained}
\acmConference[MMASIA '24]{ACM Multimedia Asia}{December 3--6, 2024}{Auckland, New Zealand}
\acmBooktitle{ACM Multimedia Asia (MMASIA '24), December 3--6, 2024, Auckland, New Zealand}\acmDOI{10.1145/3696409.3700234}
\acmISBN{979-8-4007-1273-9/24/12}

\begin{document}

\title{SITransformer: Shared Information-Guided Transformer for Extreme Multimodal Summarization}

\author{Sicheng Liu}

\affiliation{%
  \institution{The University of Sydney}
  \city{Sydney}
  \state{NSW}
  \country{Australia}
}
\email{sliu3452@uni.sydney.edu.au}

\author{Lintao Wang}
\affiliation{%
  \institution{The University of Sydney}
  \city{Sydney}
  \state{NSW}
  \country{Australia}
  }
\email{lwan3720@uni.sydney.edu.au}

\author{Xiaogang Zhu}
\affiliation{%
  \institution{The University of Adelaide}
  \city{Adelaide}
  \state{SA}
  \country{Australia}
}
\email{xiaogang.zhu@adelaide.edu.au }

\author{Xuequan Lu}
\affiliation{%
  \institution{La Trobe University}
  \city{Melbourne}
  \state{VIC}
  \country{Australia}
}
\email{b.lu@latrobe.edu.au }

\author{Zhiyong Wang}
\affiliation{%
  \institution{The University of Sydney}
  \city{Sydney}
  \state{NSW}
  \country{Australia}
}
\email{zhiyong.wang@sydney.edu.au }

\author{Kun Hu}
\authornote{Corresponding author}
\affiliation{%
  \institution{The University of Sydney}
  \city{Sydney}
  \state{NSW}
  \country{Australia}
}
\email{kun.hu@sydney.edu.au}






\renewcommand{\shortauthors}{Liu et al.}

\begin{abstract}
Extreme Multimodal Summarization with Multimodal Output (XMSMO) becomes an attractive summarization approach by integrating various types of information to create extremely concise yet informative summaries for individual modalities. Existing methods overlook the issue that multimodal data often contains more topic irrelevant information, which can mislead the model into producing inaccurate summaries especially for extremely short ones. In this paper, we propose SITransformer, a \textbf{S}hared \textbf{I}nformation-guided \textbf{T}ransformer for extreme multimodal summarization.  
It has a shared information guided pipeline which involves a cross-modal shared information extractor and a cross-modal interaction module. The extractor formulates semantically shared salient information from different modalities by devising a novel filtering process consisting of a differentiable top-k selector and a shared-information guided gating unit. As a result, the common, salient, and relevant contents across modalities are identified. 
Next, a transformer with cross-modal attentions is developed for intra- and inter-modality learning with the shared information guidance to produce the extreme summary. 
Comprehensive experiments demonstrate that SITransformer significantly enhances the summarization quality for both video and text summaries for XMSMO. Our code is publicly available at \href{https://github.com/SichengLeoLiu/MMAsia24-XMSMO}{https://github.com/SichengLeoLiu/MMAsia24-XMSMO}. 

\end{abstract}




\maketitle

\begin{figure}[t]
    \centering
    \includegraphics[width=1\linewidth]{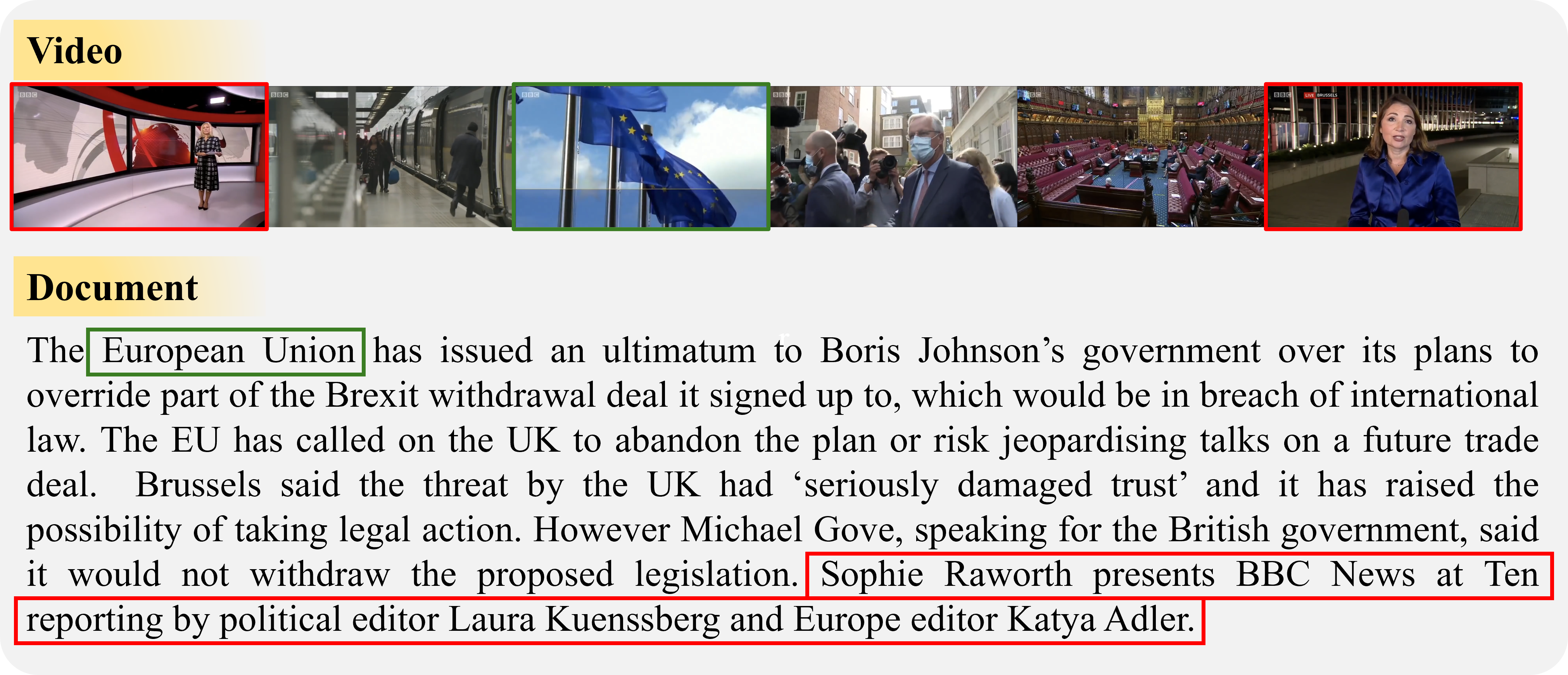}
    \caption{An example of irrelevant noises (in red box) and shared salient information (in green box).}
    \label{fig:irrelevant noise}
\end{figure}

\section{Introduction}

In the digital age, information overload is a major issue due to the rapid production of data that surpasses human processing abilities. Summarization techniques aim to distill essential information quickly, aiding in managing this overload. Traditional text summarization and video summarization methods have been extensively studied \cite{kupiec1995trainable,liu2019text,zhang2016video}. However, with the rise of multimodal content on the Internet, conventional unimodal summarization methods are increasingly inadequate. Hence, multimodal summarization with multimodal output (MSMO) \cite{zhu2018msmo}, which integrates information from multiple modalities, has emerged to address this gap. 
By incorporating diverse data sources such as videos and text documents, with advanced deep learning techniques~\cite{vaswani2017attention,hu2019vision}, it enhances summarization quality and user experience. 
To further simplify and expedite information acquisition, extreme summarization was introduced into the MSMO task \cite{tang2023tldw,tang2023topiccat}, proposing eXtreme Multimodal Summarization with Multimodal Output (XMSMO). XMSMO condenses a video-document pair into a single frame and a single sentence. This extremely concise multimodal output allows users to quickly grasp the essence and decide whether to engage further. 

Multimodal data contains rich information from various sources and formats to create more informative summary. However, the wealth of information can include noise and irrelevant details that may mislead or complicate the summarization process. 
For instance, as depicted in Figure~\ref{fig:irrelevant noise}, prolonged scenes of a host introducing news in a studio setting can be misleading if the model focuses solely on visual data. These scenes shown in red boxes do not reflect the actual news content and are typically irrelevant to the summary. 
Such challenges become more significant when generating extremely short summarization in the XMSMO task. Even minor errors can significantly reduce the quality of the summarization, underscoring the importance of accurately identifying topic-relevant information. 

To address this issue, we posit that shared information across different modalities is able to provide critical insights for comprehending the topic, thereby facilitating better utilization of the relevant information from both shared and non-shared contents for the summary. 
Thus, this paper proposes an innovative multimodal summarization method called SITransformer for the video-document XMSMO task. Overall, SITransformer consists of a \textit{cross-modal share information extractor} and a \textit{cross-modal interaction module} for the effective exploitation and utilization of shared information guidance. Specifically, the extractor aims to extract the salient information shared across modalities by a differentiable top-k selection, which ranks and selects the most salient features across modalities. The shared information is then used to control the information gate units, which determine which portions of the source features should be retained and which should be discarded, thereby filtering out features irrelevant to the topic and summarization. The cross-modal interaction module then processes the extracted relevant information using visual and textual transformers with cross-modality attentions. This enables both unimodal and cross-modal information learning to produce the final summarization.

The main contributions of this paper are as follows:
\begin{enumerate}
    \item We propose a novel summarization method, SITransformer, for the XMSMO task using cross-modal shared salient information. 
    \item We design an innovative cross-modal shared information extractor that includes a differentiable top-k selector and an information gating unit for systematically exploiting relevant multimodal information. 
    \item Comprehensive experiments on a large-scale multimodal dataset demonstrate the effectiveness of SITransformer in achieving state-of-the-art performance. 
\end{enumerate}

\section{Related Work}

\textbf{Unimodal Summarization} focuses on summarizing content from a single modality, such as text or video. In text summarization, extractive methods select key sentences from the original document \cite{liu2019fine}. Abstractive methods instead generate new sentences, often using Transformer-based models like BART \cite{lewis2019bart}. Hybrid methods, such as the Pointer-Generator Network \cite{see2017get,tang2022otextsum}, combine extraction and generation of content for the final summary. More recently, large language model-based methods for text summarization have been explored, but they are more prone to hallucination and face challenges in fine-tuning for domain-specific knowledge compared to traditional methods \cite{jin2024comprehensive}.
Video summarization involves selecting informative frames to represent the main content. Early studies relied on methods such as LSTMs to formulate temporal patterns~\cite{zhang2016video}, while recent methods used Transformer-based methods to capture frame relationships \cite{zhao2022hierarchical}. These unimodal methods establish a critical foundation for multimodal summarization.

\noindent\textbf{Multimodal Summarization} aims to improve summarization quality and user experience by summarizing information from two or more modalities. Early studies typically utilized data from other modality as the auxiliary information to facilitate the summarization regarding the target modality \cite{liu2022abstractive,zhu2023topic}. 
MSMO was first introduced by \cite{zhu2018msmo,zhu2020multimodal}, where visual attention mechanisms were designed to identify the most salient images. 
\citet{li2020vmsmo} pioneered the incorporation of video into MSMO by utilizing a recurrent neural network (RNN) to capture temporal patterns, instead of extracting a summary from a set of individual images.
\citet{zhang2022hierarchical} introduced a hierarchical cross-modality semantic correlation learning model (HCSCL) that employed a Faster R-CNN to detect visual objects. This method integrated object and text features as low-level information, and hierarchically fused sentences and scenes as high-level information. Such hierarchical fusion strategy has been used in many recent studies and proved to be an effective cross-modal information understanding method \cite{tang2023tldw, xu2023mhscnet}. 

\begin{figure*}[h!]
    \centering
    \includegraphics[width = 1\textwidth]{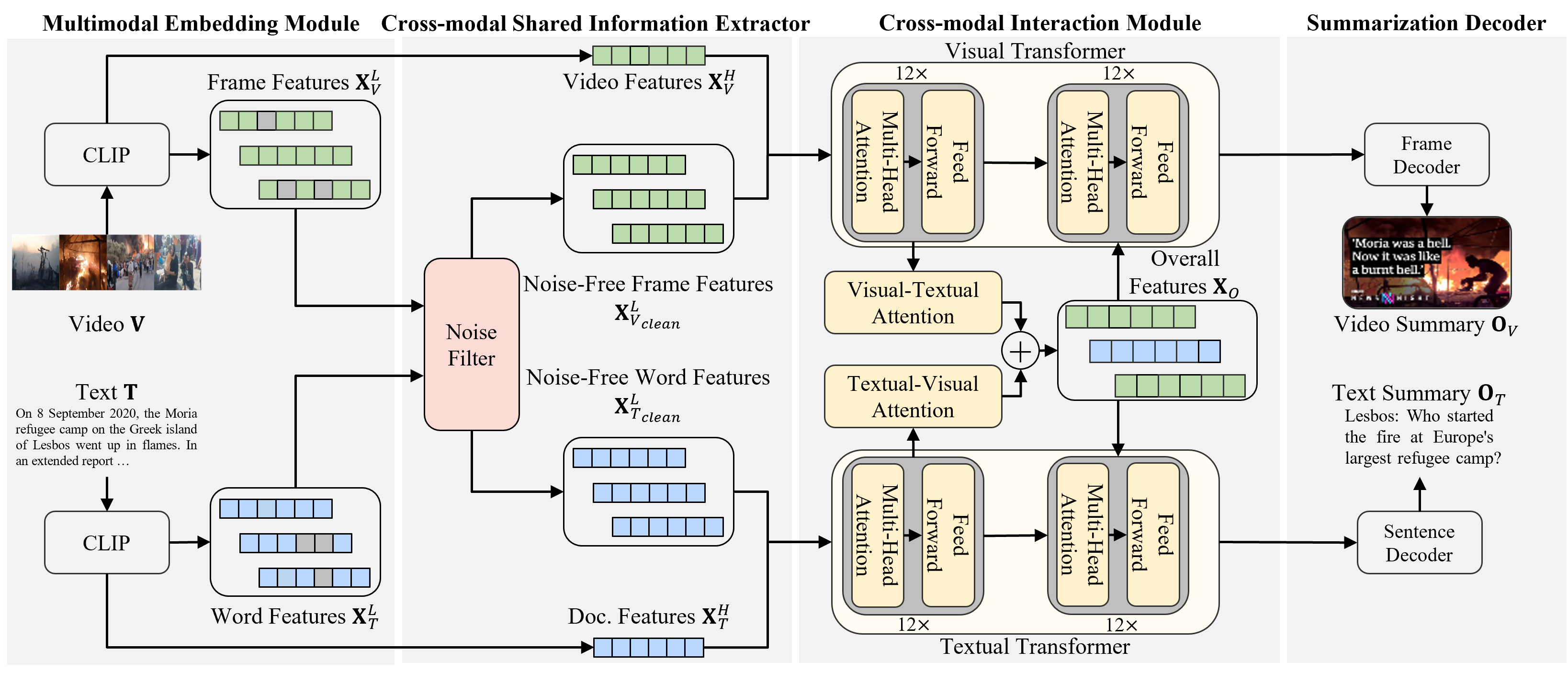}
    \caption{The overall structure of the SITransformer, which consists of a multimodal embedding module, a cross-modal shared information extractor, a cross-modal interaction module and modality-specific extreme summarization decoders.}
    \label{fig:overall structure}
\end{figure*}

\noindent\textbf{Extreme summarization} condenses original data into extremely concise content. For instance, extreme video summarization aims to distill a video into a single frame, while extreme text summarization focuses on summarizing a document into one sentence. 
\citet{tang2023tldw,tang2023topiccat} introduced the XMSMO task, which summarizes a video-document pair into a single sentence and a single frame. Their hierarchical model integrated information across modalities at sentence-scene and video-document levels. By minimizing Wasserstein distances between selected frames/texts and original videos/documents respectively, the model improves summarization coherence. These studies offer a feasible approach for extreme summarization but are often affected by irrelevant information in rich multimodal data, resulting in less promising summarization quality.

\section{Methodology}

The proposed SITransformer pipeline, shown in Figure~\ref{fig:overall structure}, includes four core stages with shared salient information guidance.

\subsection{Problem Definition}

The XMSMO task aims to summarize an input video-document pair into an extremely short one-sentence and one-cover-frame summary. 
Mathematically, each paired input can be denoted as $(\mathbf{T}, \mathbf{V})$. We represent the text document as $\mathbf{T}=\{\mathbf{t}_i | i \in [1,n]\}$ and the video as $\mathbf{V}= \{\mathbf{v}_j | j \in [1,m]\}$. Here, $\mathbf{t}_i$ denotes the $i$-th word in the document $\mathbf{T}$, and $\mathbf{v}_j$ represents the $j$-th frame in the video $\mathbf{V}$. The $m$ and $f$ indicate the corresponding lengths, respectively. The summarization outputs are defined as a single sentence $\mathbf{O}_{T} = \{\mathbf{t}^{\prime}_1, \ldots, \mathbf{t}^{\prime}_o\} \in \mathbf{T}$ and a cover-frame $\mathbf{O}_{V} = \{\mathbf{v}^{\prime}_o\} \in \mathbf{V}$ where $\mathbf{t}^{\prime}$ and $\mathbf{v}^{\prime}$ are created based on the input document and video.

\subsection{Multimodal Embedding Module}

To achieve an initial alignment between visual and textual modalities, we apply the widely used multimodal embedding model, CLIP \cite{radford2021learning}, as our latent feature encoder. 
The CLIP's multimodal latent space enables the visual and textual contents with related semantics to share similar embeddings, which is able to facilitate the cross-modal shared information exploration. 

Mathematically, for the text input $\mathbf{T}$ and video input $\mathbf{V}$,their word and frame-level low-level features, denoted as $\mathbf{X}_T^L$ and $\mathbf{X}_V^L$, respectively, can be obtained as:
\begin{equation}
\begin{aligned}
    \mathbf{X}_T^L &= \mathrm{CLIP}(\mathbf{T}) = \{\mathbf{x}_{t,i}^l | i \in [1,n]\}  \in \mathbb{R}^{n \times d}, \\
    \mathbf{X}_V^L &= \mathrm{CLIP}(\mathbf{V}) = \{\mathbf{x}_{v,j}^l | j \in [1,m]\}  \in \mathbb{R}^{m \times d},
\end{aligned}
\end{equation}
where $d$ is the latent dimension size, $n$ denotes the number of words, and $m$ denotes the number of video frames.

In addition to the low-level features (word and frame-level), high-level features (video and document-level), $\mathbf{X}_T^H$ and $\mathbf{X}_V^H$ are further considered for both document and video by using a mean pooling method across all words and frames, respectively. The video and document-level features can be obtained as:
\begin{equation}
\begin{aligned}
    \mathbf{X}_T^H &= \mathrm{pool}(\mathbf{X}_T^L) \in \mathbb{R}^{d}, \\
    \mathbf{X}_V^H &= \mathrm{pool}(\mathbf{X}_V^L) \in \mathbb{R}^{d}.
\end{aligned}
\end{equation}

\subsection{Cross-Modal Shared Information Extractor} \label{sec: noise removal}

The low and high-level features contain rich information relevant to the major topic. However, they often include significant irrelevant information, making precise summarization challenging.
To reduce such negative impact, a cross-modal shared information extractor with noise filtering mechanism is devised. 
The core idea is to leverage shared salient information, presenting in both modalities, to regulate information gates and filter out irrelevant information from the shared and non-shared feature embeddings accordingly.

\begin{figure}[t]
    \centering
    \includegraphics[width = 1\linewidth]{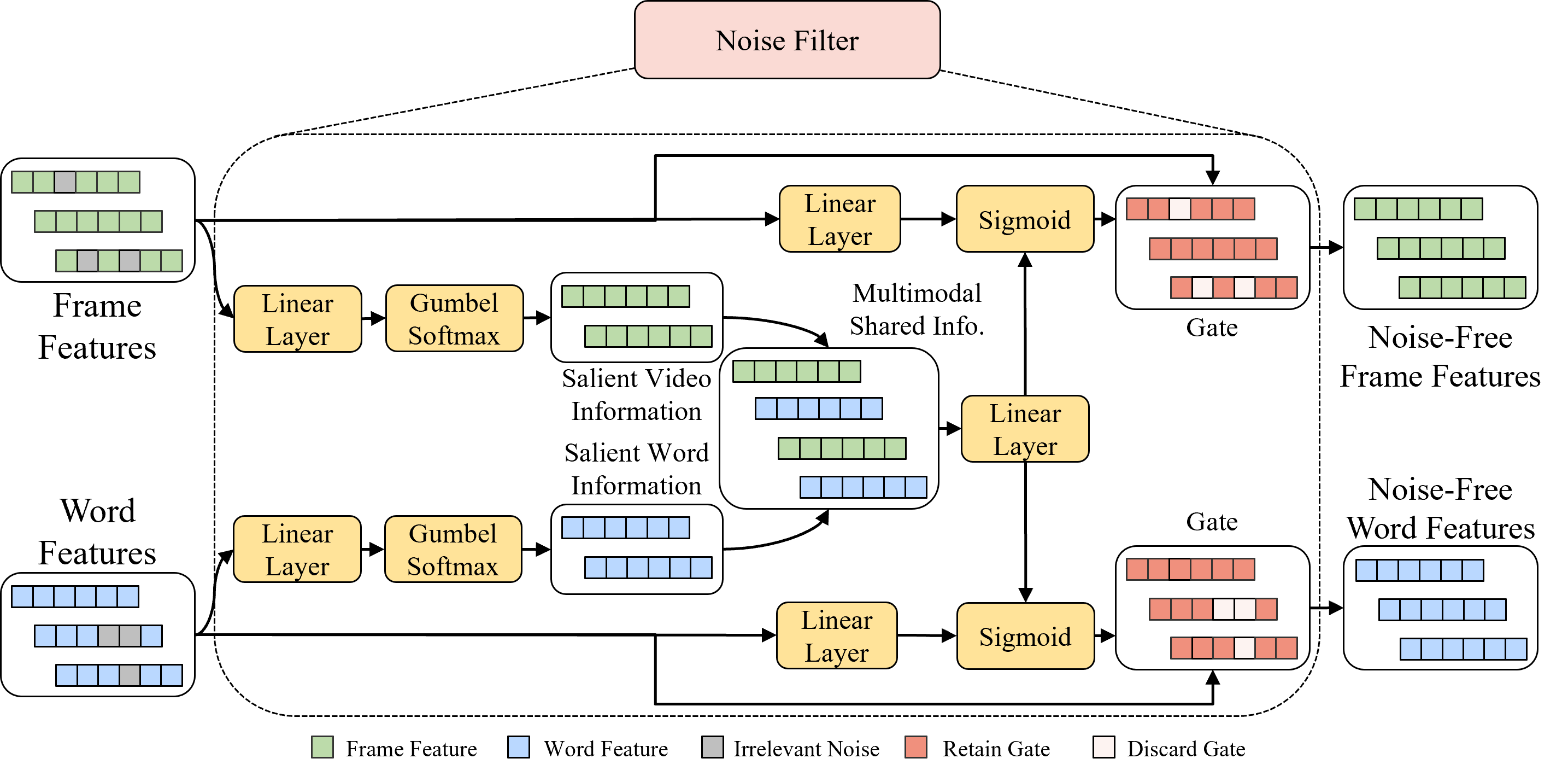}
    \caption{Noise Filter with Differentiable Top-k (NFDT). 
    }
    \label{fig:noise filter topk}
\end{figure}

The noise filtering mechanism is designed with a Differentiable Top-k (NFDT) \cite{cordonnier2021differentiable} strategy as depicted in Figure~\ref{fig:noise filter topk}.
Specifically, linear layers are first adopted to obtain word/frame salient scores:
\begin{equation}
\begin{aligned}
    \mathbf{s}_T &= \mathbf{W}_1 \mathbf{X}_T^L + \mathbf{b}_1 \in \mathbb{R}^{ n},\\
    \mathbf{s}_V &= \mathbf{W}_2 \mathbf{X}_V^L + \mathbf{b}_2 \in \mathbb{R}^{ m},
\end{aligned}
\end{equation}
where $\mathbf{S}_T=\{s_{t,i} | i \in [1,n]\}$ and $\mathbf{S}_V=\{s_{v,m} | i \in [1,f]\}$ are the scores for each word and frame; $\mathbf{W}_1$, $\mathbf{W}_2$, $\mathbf{b}_1$ and $\mathbf{b}_2$ are learnable weight and bias, respectively. 

Next, Gumbel softmax functions \cite{jang2016categorical} are utilized to standardize the score between 0 and 1. Compared to the regular softmax function, the Gumbel softmax function is differentiable by integrating noise $g$ sampled from Gumbel distribution by: 
\begin{equation}
    g=-\mathrm{log}(-\mathrm{log}(u)) \in \mathbb{R}, 
\end{equation}
where $u$ is a random variable uniformly distributed on $[0,1]$. 
Gumbel softmax is then calculated as follows:
\begin{equation}
\begin{aligned}
    \text{score}_{t,i} &= \frac{\mathrm{exp}((s_{t,i} + g_i)/\tau)}{\sum_{i=1}^n \mathrm{exp}((s_{t,i} + g_i)/\tau)} \in \mathbb{R}, \\
    \text{score}_{v,j} &= \frac{\mathrm{exp}((s_{v,j} + g_j)/\tau)}{\sum_{j=1}^m \mathrm{exp}((s_{v,j} + g_j)/\tau)} \in \mathbb{R},
\end{aligned}
\end{equation}
where the $\text{score}_{t,i}$ and $\text{score}_{v,j}$ are the final scores for the $i$-th word and the $j$-th frame, respectively; $\tau$ is a temperature parameter that controls the smoothness. 

We can obtain the score list $\mathbf{S}_{T^\prime} = \{score_{t,i} | i \in [1,n]\}$ and $\mathbf{S}_{V^\prime} = \{score_{v,j} | j \in [1,m]\}$ for all individual texts and frames. The top-k features with the highest estimated scores are selected to construct the potential multimodal shared salient information $\mathbf{X}_S$:
\begin{equation}
\begin{aligned}
    \mathbf{X}_S &= \mathrm{Top}_k(\mathbf{S}_{T^\prime}) \cup \mathrm{Top}_k(\mathbf{S}_{V^\prime}) \\
    &= \{\mathbf{x}^l_{t,n_1}, \ldots, \mathbf{x}^l_{t, n_k}\} \cup \{\mathbf{x}^l_{v,m}, \ldots, \mathbf{x}^l_{v, m_k}\} \in \mathbb{R}^{2|k| \times d}, 
\end{aligned}
\end{equation}
where $k$ is a hyper-parameter that controls the amount of selected multimodal shared salient information. 

Overall, the multimodal shared salient information $\mathbf{X}_S$ contains $k$ image and word features simultaneously. 
It will be used to concatenate with word and frame features respectively to provide shared information guidance to the irrelevant information filtering process. The shared information is first pooled as: 
\begin{equation}
    \mathbf{x}_S = \mathrm{pool}(\mathbf{X}_S) \in \mathbb{R}^{  d}.
\end{equation}

The pooled information is then used to control an novel information gating unit. Specifically, if the gate vector is 0, the feature is considered irrelevant and discarded; otherwise, it is retained. The gate vector is defined as:
\begin{equation}
\begin{aligned}
    \mathbf{G}_T &= \mathrm{sigmoid}(\mathbf{W}_3(\mathbf{x}_S \oplus \mathbf{X}_T^L)+\mathbf{b}_3) \in \mathbb{R}^{n \times d}, \\
    \mathbf{G}_V &= \mathrm{sigmoid}(\mathbf{W}_4(\mathbf{x}_S \oplus \mathbf{X}_V^L)+\mathbf{b}_4) \in \mathbb{R}^{m \times d},
\end{aligned}
\end{equation}
where $\mathbf{G}_T$ and $\mathbf{G}_V$ are the gate matrices for text and frames, respectively. $\mathbf{W}_3$, $\mathbf{W}_4$, $\mathbf{b}_3$ and $\mathbf{b}_4$ are learnable parameters.  Here, $(\mathbf{x}_S \oplus \mathbf{X}_T^L)$ and $(\mathbf{x}_S \oplus \mathbf{X}_V^L)$ represent the broadcasting of $\mathbf{x}_S$ and its concatenation to the beginning of each row of $\mathbf{X}_T^L$ and $\mathbf{X}_V^L$, respectively.

Finally, we filter out irrelevant information with the shared salient information guided gate by dot multiplying the original text and frame features with the corresponding gate vector.

\begin{equation}
\begin{aligned}
    \mathbf{X}_{T_{clean}}^L &= \mathbf{X}_T^L \cdot \mathbf{G}_T \in \mathbb{R}^{n \times d}, \\
    \mathbf{X}_{V_{clean}}^L &= \mathbf{X}_V^L \cdot \mathbf{G}_V \in \mathbb{R}^{ m \times d},
\end{aligned}
\end{equation}
where $\mathbf{X}_{T_{clean}}^L$ and $ \mathbf{X}_{V_{clean}}^L$ are the noise-free word and frame features, respectively.

\subsection{Multi-Level Unimodal Representations}

Based on the salient low-level features, unimodal representations can be finally obtained jointly with the high-level features. 
For the input document, document- and word-level features are concatenated first. Next, a Transformer is utilized to introduce a video-level understanding to all frame-level features: 
\begin{equation}
\begin{aligned}
    \mathbf{X}_T^{\prime} &= \mathbf{X}_{T_{clean}}^L \oplus \mathbf{X}_T^H \in \mathbb{R}^{ (n+1) \times d},\\
    \mathbf{X}_T &= \mathrm{Transformer}(\mathbf{X}_T^{\prime}) \in \mathbb{R}^{ n \times d}.
\end{aligned}
\end{equation}
Note that we discard the high-level information token in the output as it has already exchanged information with the low-level information. Similarly, we can process the frame-level features as:
\begin{equation}
\begin{aligned}
    \mathbf{X}_V^{\prime} &= \mathbf{X}_{V_{clean}}^L \oplus \mathbf{X}_V^H \in \mathbb{R}^{ (m+1) \times d},\\
    \mathbf{X}_V &= \mathrm{Transformer}(\mathbf{X}_V^{\prime}) \in \mathbb{R}^{ m \times d}.
\end{aligned}
\end{equation}

\subsection{Cross-Modal Interaction Module}

To facilitate the multimodal learning and explore the cross-modal interaction, a cross-attention based mechanism is adopted. We formulate the attention scores from (1) video to document and (2) document to video as: 
\begin{equation}
\begin{aligned}
    \mathbf{A}_{V2T} &= \mathrm{Attention}(\mathbf{Q}=\mathbf{X}_V, \mathbf{K}=\mathbf{X}_T, \mathbf{V}=\mathbf{X}_T)  \in \mathbb{R}^{ m \times d}, \\
    \mathbf{A}_{T2V} &=  \mathrm{Attention}(\mathbf{Q}=\mathbf{X}_T, \mathbf{K}=\mathbf{X}_V, \mathbf{V}=\mathbf{X}_V) \in \mathbb{R}^{ n \times d}.
\end{aligned}
\end{equation}
This mechanism enables effective information interaction between video and document, enhancing multimodal understanding. 
Next, a multimodal embedding $\mathbf{X}_O$ can be obtained by:
\begin{equation}
    \mathbf{X}_O = \mathbf{W}_5(\mathbf{A}_{V2T} \oplus \mathbf{A}_{T2V}) + \mathbf{b}_5 \in \mathbb{R}^{(m+n) \times d}.
\end{equation}

To produce summaries for each modality with the multimodal guidance, we leverage a Transformer to inject the guidance: 
\begin{equation}
\begin{aligned}
    \mathbf{X}_{TO} = \mathrm{Transformer}(\mathbf{X}_O \oplus \mathbf{X}_T) \in \mathbb{R}^{ n \times d}, \\
    \mathbf{X}_{VO} = \mathrm{Transformer}(\mathbf{X}_O \oplus \mathbf{X}_V) \in \mathbb{R}^{ m \times d},
\end{aligned}
\end{equation}
where the $\mathbf{X}_{TO}$ and $\mathbf{X}_{VO}$ are the multimodal aware word features and frame features, respectively. Note that we discard the multimodal relevant tokens for the outputs. 

\subsection{Extreme Summarization Decoder}
Based on the multimodal features $\mathbf{X}_{TO}$ and $\mathbf{X}_{VO}$, the one-sentence-summary and cover-frame are created by the corresponding extreme summarization decoders.
For text summary, a linear layer is used to compute the scores for the words in the document. 
Top-$k'$ words are selected from the input document for the generation of the one-sentence-summary $\mathbf{O}_T$: 
\begin{equation}
    \mathbf{O}_T = \mathrm{Top}_{k'}(\mathrm{Softmax}(\mathrm{Linear}(\mathbf{X}_{TO}))),
\end{equation}
where $k'$ is the maximum number of words in summary.
The selected $k'$ words are ranked by their scores from the word-wise linear layer with the softmax activation, allowing the summary sentence to be constructed by ordering these words accordingly.

The cover frame selection is treated as a classification task. We use a linear layer with softmax activation to score the frames and identify the one $\mathbf{O}_V$ with the maximum score as the cover frame:
\begin{equation}
    \mathbf{O}_V = \mathop{\mathrm{argmax}}\limits_{j}( \mathrm{Softmax}(\mathrm{Linear}(\mathbf{X}_{VO}))).
\end{equation}

\subsection{Loss and Optimization}

The proposed method is trained in an unsupervised manner.  We employ the Wasserstein distance to assess the fidelity of generated summaries, a method well-established in prior research \cite{arjovsky2017wasserstein, tang2023tldw, tang2023topiccat}. This metric helps measure summarization quality as distributional differences. The calculation of Wasserstein distance between two distributions $p$ and $q$ is:

\begin{equation}
\begin{gathered}
W\left(p, q\right)=\underset{\mathbf{\Gamma}\left(p, q\right)}{\operatorname{argmin}} \sum_{i, j} \gamma_{i j}\left(p_i, q_j\right) c_{i j}, \\
\text { s.t. } \sum_{j=1}^d \gamma_{i j}\left(p_i, q_j\right)=p_i, \sum_{i=1}^d \gamma_{i j}\left(p_i, q_j\right)=q_j, \\
\gamma_{i j}\left(p_i, q_j\right) \geq 0, \forall i, j .
\end{gathered}
\end{equation}
where $c_{i,j}$ is the Euclidean distance based cost between $p$ and $q$, $d$ is the number of features. 
$\gamma_{ij}$ represents the mass moved from $p_i$ to $q_j$, where $p_i$ and $q_j$ represent the discrete points in the feature spaces associated with the distributions $p$ and $q$, respectively.

Following \citet{tang2023tldw}, we adopt three types of Wasserstein distance losses to represent the optimal transport from predicted distribution to desired distribution as follows: 
\begin{equation}
    \mathcal{L}_T = W(p_T, q_T), \mathcal{L}_V = W(p_V, q_V), \mathcal{L}_O = W(p_O, q_O).
\end{equation}
$\mathcal{L}_T$ calculates the distributional distance between the one-sentence-summary and the input document, where $p_t$ and $q_t$ are probability mass functions of document feature $\mathbf{X}^H_T$ and summary feature $\mathbf{X}_{TO}$. Similarly, $\mathcal{L}_V$ represents the distance between the cover-frame summary and the input video probability mass function.  $\mathcal{L}_O$ ensures the output summary is consistent between sentences and frames.

In addition to summary accuracy, we use Syntactic Log-Odds Ratio (SLOR) \cite{kann2018sentence} to ensure the fluency of the textual summary as:
\begin{equation}
    \mathcal{L}_f = \frac{1}{|\mathbf{\mathbf{O}_T}|}(\mathrm{ln}(P_{LM}(\mathbf{O}_T))-\mathrm{ln}(P_{U}(\mathbf{\mathbf{O}_T}))),
\end{equation}
where $P_{LM}(\mathbf{O}_T)$ is the probability of the one-sentence assigned by a large language model (GPT-2 in practice) and $P_U(\mathbf{O}_T)$ is the unigram probability.
To this end, the overall loss can be calculated as:
\begin{equation}
    \mathcal{L} = \lambda_T\mathcal{L}_T + \lambda_V\mathcal{L}_V + \lambda_O\mathcal{L}_O + \lambda_f\mathcal{L}_f,
\end{equation}
where $\lambda_T$, $\lambda_V$, $\lambda_S$, $\lambda_f$ are hyper-parameters to control loss weight.

\section{Experiments \& Discussions}

\subsection{Experimental Settings}
\noindent\textbf{Dataset}. We adopted the multimodal news dataset proposed in~\citet{tang2023tldw}. It contains 4,891 BBC News video-document pairs. The average video duration is 345.5 seconds and the average length of documents is 498 words.
We extracted one frame every 120 frames, and a total of 60 frames were extracted as the video inputs. All video frames were resized to 640*360.

\noindent\textbf{Methods for Comparison}. We compared our SITransformer with the state-of-the-art unimodal extreme summarization methods \cite{see2017get,qi2020prophetnet,zhang2020pegasus,apostolidis2021combining,apostolidis2022summarizing}, MSMO \cite{li2020vmsmo,radford2021learning} and XMSMO methods \cite{tang2023topiccat,tang2023tldw}. 

\noindent\textbf{Evaluation Metrics}. Following existing practice \cite{tang2023tldw}, for the quantitative evaluation of video summarization, intersection over union (IoU) \cite{IoU} and frame accuracy (FA) \cite{frameaccuracy} are adapted. In addition, ROUGE \cite{rouge} metric is adopted for the text summarization.

\noindent\textbf{Implementation Details}. We used AdamW \cite{loshchilov2017decoupled} as the optimizer with a learning rate of 0.001 and a 12-layer Transformer with 16 attention heads for both unimodal data processing and cross-modal interaction. The hidden size of the Transformer was set to 512. The temperature parameter $\tau$ used in NFDT was set to 0.5. 

\subsection{Quantitative Analysis}

\begin{table}[t!]
    \centering
    \begin{tabular}{@{\extracolsep{2pt}} p{3cm}  ccccc @{}}
        \hline  
        \multirow{2}{*}{Methods} & \multicolumn{2}{c}{Visual $\uparrow$} & \multicolumn{3}{c}{Textual $\uparrow$}  \\
        \cline { 2 - 3 } \cline { 4 - 6 }& FA & IoU & R-1 & R-2 & R-L \\
        \hline
        PG \cite{see2017get} & - & - & 2.43 & 0.08 & 2.25 \\
        ProphetNet \cite{qi2020prophetnet} & - & - & 3.77 & \underline{0.09} & 3.56 \\
        PEGASUS-XSUM \cite{zhang2020pegasus} & - & - & 4.36 & \textbf{0.12} & 4.00 \\
        \hline
        CA-SUM \cite{apostolidis2022summarizing} & 0.57 & 0.69 & - & - & - \\
        ARL \cite{apostolidis2021combining} & 0.59& 0.68 & - & - & - \\
        \hline
        VMSMO \cite{li2020vmsmo} & 0.57 & 0.69 & divg & divg & divg \\
        CLIP \cite{radford2021learning} & 0.58 & 0.68 & 3.35 & 0.05 & 3.14 \\
        TLDW \cite{tang2023tldw} & 0.57 & 0.68 & 4.64 & 0.07 & 4.33 \\
        TopicCAT \cite{tang2023topiccat} & 0.60 & 0.69 & 4.54 & 0.08 & 4.21 \\
        \hline
        Ours & \textbf{0.64} & \textbf{0.79} & \textbf{6.37} & \underline{0.09} & \textbf{6.11} \\
        \hline
        w/o Shared info sel. & 0.58 & 0.7 & 5.24& 0.06& 5.10\\
        w/o Gumbel-softmax & \underline{0.61} & \underline{0.73} & \underline{5.83}& 0.07& \underline{5.38}\\
        w/o Shared info gate & 0.59 & 0.73 & 5.02 & 0.01 & 4.54 \\
        w/ Random selector & 0.59 & 0.7 & 4.37 & 0.02 & 3.93 \\
        w/ NFCS & \underline{0.61} & 0.65 & 5.67 & 0.06 & 5.23 \\
        \hline
    \end{tabular}
    \caption{Quantitative evaluation on BBC news dataset \cite{tang2023tldw}.} 
        \label{tab:quantitative analysis}
\end{table}

Table~\ref{tab:quantitative analysis} lists the quantitative results. Overall, our  SITransformer outperforms the state-of-the-art methods regarding both extreme visual and textual summarization metrics. 
For the visual summarization, SITransformer achieves a FA of 0.64 and an IoU of 0.79. The increase in FA and IoU highlights the the SITransformer's efficacy in identifying the most informative frame. Moreover, SITransformer achieves 6.37, 0.09, 6.11 on ROUGE-1, ROUGE-2 and ROUGE-L, respectively. The improvements in ROUGE metrics indicate the SITransformer's enhanced ability to capture keywords and create more fluent sentences. 
Compared to unimodal extreme summarization methods, SITransformer shows better performance in both text and video summarization, highlighting the advantages of using multimodal data. Moreover, our approach outperforms all existing (X)MSMO methods, demonstrating the effectiveness of leveraging shared salient information for the summarization process.

\subsection{Qualitative Analysis}

\begin{figure*}[ht!]
    \centering
    \includegraphics[width=0.95\textwidth]{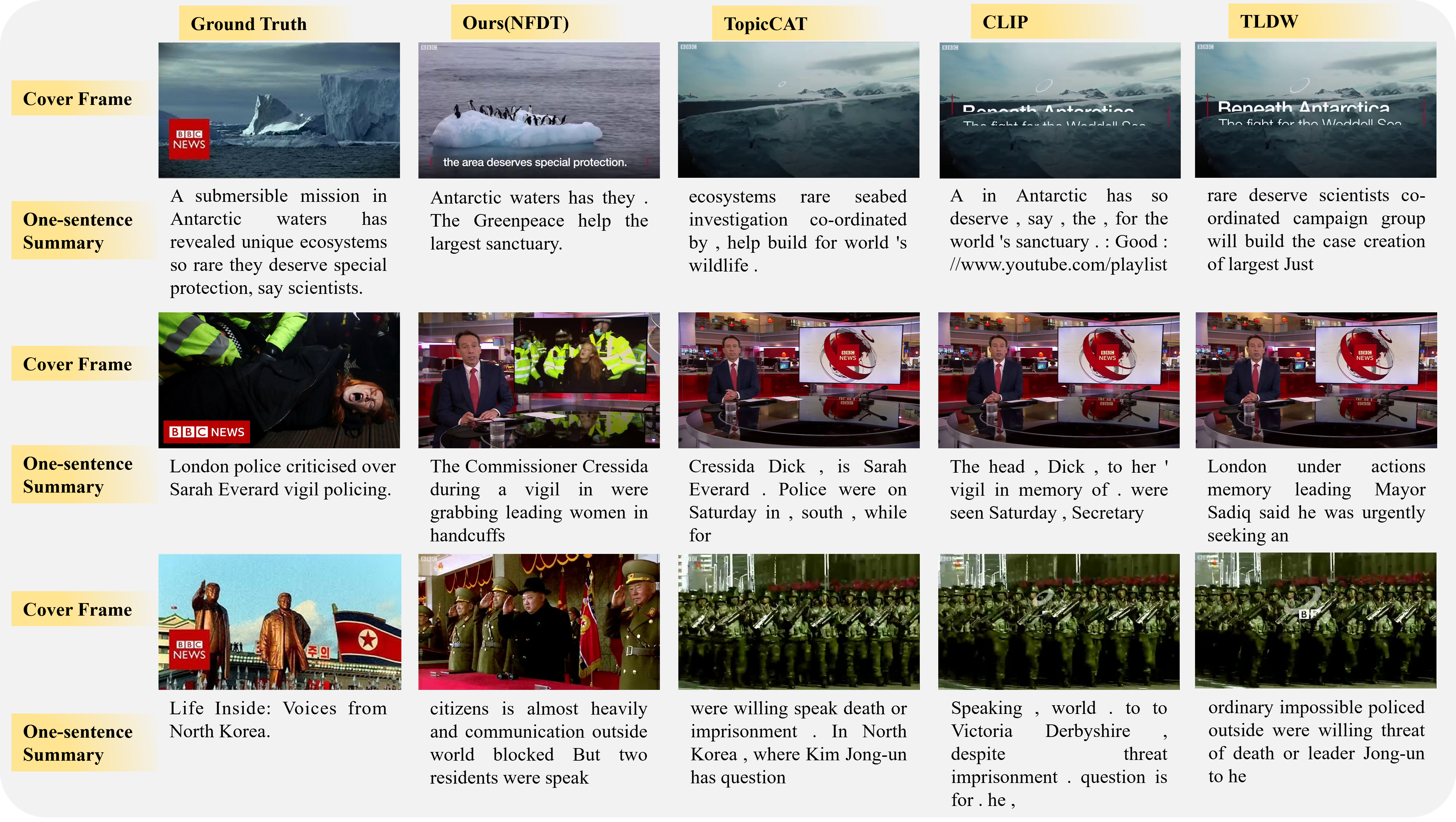}
    \caption{Qualitative examples by our NFDT and the state-of-the-art methods: TopicCAT \cite{tang2023topiccat}, TLDW \cite{tang2023tldw} and CLIP \cite{radford2021learning}.}
    \label{fig:visual summary}
\end{figure*}

We compare a number of summaries produced by our method, TopicCAT \cite{tang2023topiccat}, CLIP \cite{radford2021learning}, TLDW \cite{tang2023tldw} and the ground truth in Figure~\ref{fig:visual summary}. 
From the visual perspective, the cover frame selected by SITransformer is more consistent with the ground truth, while the selection of baselines will be affected greatly by irrelevant information. In the three examples, SITransformer selects cover frames with semantics similar to the ground truth, such as a Koala, while the baseline struggles to distinguish key information from irrelevant details and mistakenly selects irrelevant frames, like the news narrator, as cover frames.

As for the text summary, SITransformer is able to better capture the essence of the documents corresponding to the ground truths. However, all methods could be struggling in language fluency and grammatical errors, leading to room of improvement. Such problems are common for extractive summarization methods as they prioritize capturing keywords over language fluency\cite{tang2023topiccat}. Consequently, extractive summaries emphasize topic coverage, leading to a trade-off between fluency and coverage.

\subsection{Ablation Study}

We conducted ablation studies to verify the effectiveness of each proposed component. The experimental results are listed in Table \ref{tab:quantitative analysis}. 
Firstly, we removed the shared information selection process in obtaining the multimodal shared salient information. The significantly degraded performance indicates the importance of extracting shared salient information to guide the summarization. 
We also assess the impacts of Gumbel-softmax by replacing with standard softmax, and evaluate shared information guided gate design by its removal in our proposed NFDT. The results demonstrate their positive contributions to enhancing the extraction of shared salient information.  
Lastly, to verify the effectiveness of the proposed noise filter design, we replace our NFDT noise filter with two different types of filters, namely random feature selector and noise filter with cosine similarity (NFCS). The former filters shared salient features by random selection while the later applies cosine similarity based ranking for salient information identification. 
Compared to our best model, both visual and textual summary metrics degrade, demonstrating the performance of our proposed novel noise filter.

\begin{figure}[t]
    \centering
    \subfigure{
        \begin{minipage}[b]{0.497\linewidth}
        \includegraphics[width = 1\linewidth]{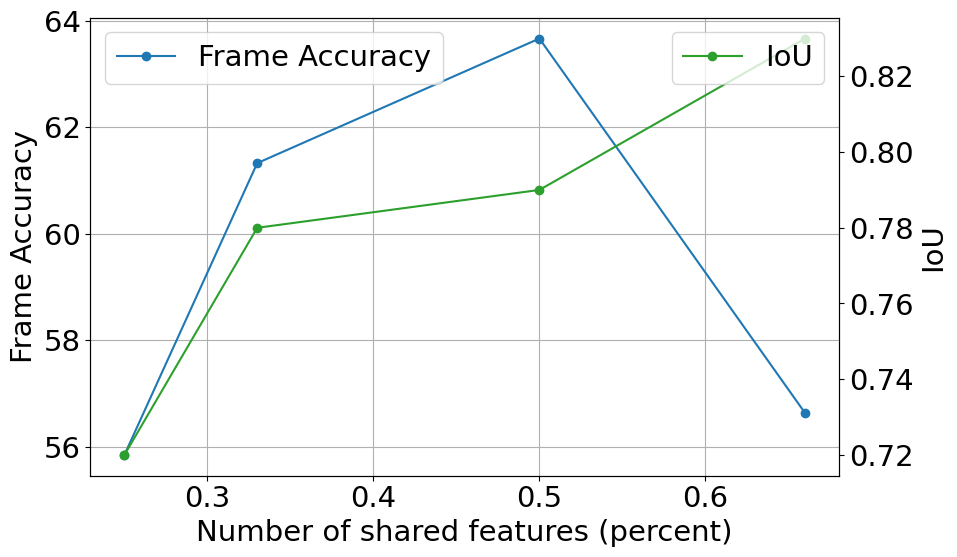}\label{fig:result curve1}
        \end{minipage}
    }
    \subfigure{
        \begin{minipage}[b]{0.455\linewidth}
        \includegraphics[width = 1\linewidth]{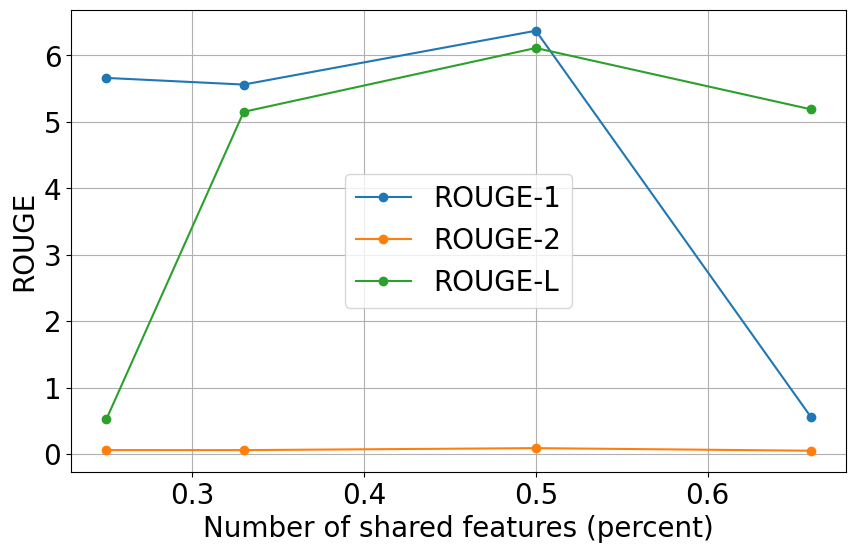}\label{fig:result curve2}
        \end{minipage}
    }
    \caption{Impacts of shared salient information amount} 
    \label{fig: result curve}
\end{figure}

\subsection{Impacts of Shared Information Amount}
The amount of shared salient information to extract is an important hyper-parameters.
We visualize its impacts during the multimodal shared information extraction process in Figure ~\ref{fig: result curve}. 
It shows that retaining 50\% of shared information improves most metrics except IoU. 
Selecting either an excessive or insufficient number of features can adversely affect performance. Choosing too many shared salient features may result in the inclusion of undesirable, irrelevant information, leading to the loss of valuable data. Conversely, using too few shared salient features may fail to adequately control the noise filter, also causing performance degradation.

\subsection{Limitations and Future Work}

SITransformer can produce sub-optimal text summarization for some scenarios, exhibiting grammatical errors and structural inconsistencies. One potential solution is to leverage large language models (LLMs) such as BLIP2 \cite{li2023blip} and MiniGPT5 \cite{zheng2023minigpt} to further refine the fluency and grammar accuracy of the results. Furthermore, future work could explore methods that generate cover frames based on user preferences and customization, using diffusion methods~\cite{rombach2022high}, which further improve the flexibility of XMSMO. 

\section{Conclusion}

This paper presents SITransformer for XMSMO, which leverages cross-modal shared information to guide the summarization process. 
A cross-modal shared information extractor is devised to identify shared salient information and a transformer based cross-modal interaction module facilitates the cross-modality learning. Our extensive experiments have demonstrated the effectiveness of our proposed method.

\bibliographystyle{ACM-Reference-Format}

\appendix

\end{document}